\documentclass[10pt,twocolumn,letterpaper]{article}

\usepackage{cvpr}
\usepackage{times}
\usepackage{epsfig}
\usepackage{graphicx}
\usepackage{amsmath}
\usepackage{amssymb}
\usepackage{algorithm}
\usepackage{subfigure}
\usepackage{bm}

\usepackage[breaklinks=true,bookmarks=false]{hyperref}
\cvprfinalcopy 

\def\cvprPaperID{2255} 

\ifcvprfinal\fi
\begin{document}

\title{Noisy Softmax: Improving the Generalization Ability of DCNN via Postponing the Early Softmax Saturation}
\author{Binghui Chen$^{1}$, Weihong Deng$^{1}$, Junping Du$^{2}$\\
$^{1}$School of Information and Communication Engineering, Beijing University of Posts and Telecommunications,\\
$^{2}$School of Computer Science, Beijing University of Posts and Telecommunications, Beijing China.\\
{\tt\small chenbinghui@bupt.edu.cn, whdeng@bupt.edu.cn, junpingd@bupt.edu.cn}
}

\maketitle
\thispagestyle{empty}

\begin{abstract}
Over the past few years, softmax and SGD have become a commonly used component and the default training strategy in CNN frameworks, respectively. However, when optimizing CNNs with SGD, the saturation behavior behind softmax always gives us an illusion of training well and then is omitted. In this paper, we first emphasize that the \textbf{early saturation} behavior of softmax will impede the exploration of SGD, which sometimes is a reason for model converging at a bad local-minima, then propose \textbf{Noisy Softmax} to mitigating this early saturation issue by injecting annealed noise in softmax during each iteration. This operation based on noise injection aims at postponing the early saturation and further bringing continuous gradients propagation so as to significantly encourage SGD solver to be more exploratory and help to find a better local-minima. This paper empirically verifies the superiority of the early softmax desaturation, and our method indeed improves the generalization ability of CNN model by regularization. We experimentally find that this early desaturation helps optimization in many tasks, yielding state-of-the-art or competitive results on several popular benchmark datasets.
\end{abstract}

\section{Introduction}
Recently, deep convolutional neural networks (DCNNs) have taken the computer vision field by storm, significantly improving the state-of-the-art performances in many visual tasks, such as face recognition
~\cite{Sun2014Deep, Sun2014Deeply, Parkhi2015Deep, Schroff2015FaceNet}, large-scale image classification ~\cite{Krizhevsky2012ImageNet, Simonyan2015Very, Szegedy2015Going, He2015Deep, He2016Identity}, and fine-grained object classification~\cite{Lin2015Bilinear, Huang2015Part, Krause2015Fine, Wei2016Mask}.
Meanwhile, softmax layer and the training strategy of SGD together with back-propagation (BP) become the default components, and are generally applied in most of the aforementioned works.

\begin{figure}[t]
  \centering
  \begin{minipage}{1\linewidth}
  \includegraphics[width=1\linewidth]{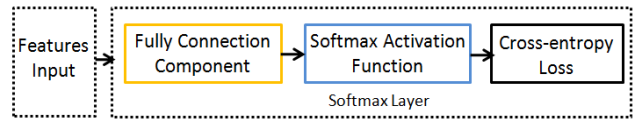}\\
    \vspace{-1em}
  \caption{Decomposition of typical softmax layer in DCNN. It can be rewritten into three parts: fully connection component, softmax activation and cross-entropy loss.}\label{fig:decomposition}
  \end{minipage}
  \vspace{-2em}
\end{figure}

It is widely observed that when optimizing with SGD and BP, the smooth and free gradients propagation is crucial to improve the training of DCNNs. For example, replacing sigmoid activation function with the piecewise-linear activation functions such as ReLU and PReLU~\cite{He2015Delving} handles the problem of gradients vanishing caused by sigmoid saturation and, allows the training of much deeper networks. While, it is interesting that softmax activation function (illustrated in Figure~\ref{fig:decomposition}) is implicitly like sigmoid function due to their similar formulation (shown in Sec.~\ref{sec:individual}), and has the saturation behavior as well when its input is large. However, many take the softmax activation for granted and the problem behind its saturation behavior is omitted as a result of illusion of performance improvements based on DCNNs.

In standard SGD, the saturation behavior of softmax turns up when its output is very close to the ground truth, certainly it is our goal of model training. However, in some ways, it is a barrier to improve the generalization ability of CNNs especially when it shows up early (inopportune). Concretely, for one instance input, it will early stop contributing gradients to BP when its softmax output is prematurely saturated, yielding short-lived gradients propagation in history which is not enough for robust learning. And in this case, the learning process with SGD and BP hardly explore more due to poor gradients propagation and parameters update. We define this saturation behavior as \emph{individual saturation} and the corresponding individual as saturated one. As the training going, the number of non-saturated contributing training samples gradually decreases and the robust learning of network will be impeded. It sometimes is a reason for algorithm going to bad local-minima\footnote{For simplicity, we use local or 'global' minima to represent a neighbouring region not a single point.} and being difficult to escape. Furthermore, the problem of over-fitting turns up. To this end, we need to give SGD chances to explore more parts of parameter space and the early individual saturation is undesired.


In this paper, we propose Noisy Softmax, a novel technique of early softmax desaturation, to address the aforementioned issue. This is mainly achieved by injecting annealed noise directly into softmax activations during each iteration. In another word, Noisy Softmax allows SGD to escape from a bad local-minima and explore more by postponing the early individual saturation. Furthermore, it improves the generalization ability of system by reducing over-fitting as a direct consequence of more exploration. The main contributions of this work are summarized as
follows:
\vspace{-0.5em}
\begin{itemize}
  \item We provide an insight of softmax saturation, interpreted as \emph{individual saturation}, that early individual saturation produces short-lived gradients propagation which is poor for robust exploration of SGD and further causes over-fitting unintentionally. 
  \vspace{-0.5em}
  \item We propose \textbf{Noisy Softmax} to aim at producing rich and continuous gradients propagation by injecting annealed noise into softmax activations. It allows the 'global' convergence of SGD solver and aids generalization by reducing over-fitting. To our knowledge, it is the first attempt to address the early saturation behavior of softmax by adding noise.
  \vspace{-0.5em}
  \item Noisy Softamx can be easily performed as a drop-in replacement for standard softmax and optimized with standard SGD. It can be also applied in other performance-improving techniques, such as neural activation functions and network architectures.
  \vspace{-0.5em}
  \item Extensive experiments have been performed on several datasets, including MNIST~\cite{LecunThe}, CIFAR10/100~\cite{Krizhevsky2012Learning}, LFW~\cite{Huang2008Labeled}, FGLFW~\cite{Nanhai2016Fine} and YTF~\cite{Wolf2011Face}. The impressive results demonstrate the effectiveness of Noisy Softmax.
\end{itemize}
\vspace{-0.5em}
\section{Related Work}

Many promising techniques have been developed, such as novel network structures~\cite{Lin2013Network, He2016Identity, Srivastava2015Highway}, non-linear activation functions~\cite{He2015Delving, Goodfellow2013Maxout, Glorot2010Deep, Shang2016Understanding}, pooling strategies~\cite{He2015Deep, Graham2014Fractional, Zeiler2013Stochastic} and objective loss functions~\cite{Sun2014Deep, Schroff2015FaceNet}, \etc.

These approaches are mostly optimized with SGD and back-propagation. In standard SGD, we use the chain rule to compute and propagate the gradients. Thus, any saturation behavior of neuron units or layer components\footnote{The saturation of layer refers to gradients vanishing at a certain layer in back-propagation.} are undesired because the training of deeper framework attributes to smooth and free flow of gradients information. The early solution is to replace sigmoid function with non-linear piecewise activation function
~\cite{Krizhevsky2012ImageNet}. This is neuron desaturation.
Skip connections between different layers exponentially expands the paths of propagation~\cite{Srivastava2015Highway, He2015Deep, He2016Identity, Huang2016Deep, Huang2016Densely}. These belong to layer desaturation, since the forward and backward information can be directly propagated from one layer to any other layer without gradients vanishing. In contrast, only the early saturation behavior is harmful instead of all of them, we focus on the early desaturation of softmax, which hasn't been investigated, and we achieve this by injecting noise explicitly into softmax activations.

There are some other works that are related to noise injection. Adding noise to ReLU has been developed to encourage components to explore more in Boltzmann machines and feed-forward networks~\cite{by2015Rectified, Bengio2013Estimating}. Adding noise to sigmoid provides possibilities of training with much wider family of activation functions than previous~\cite{Gulcehre2016Noisy}. Adding weight noise~\cite{Mark2000A}, adaptive weight noise~\cite{Graves2011Practical, Blundell2015Weight} and gradients noise~\cite{Neelakantan2015Adding} also improve the learning. Adding annealed noise can help the solver escape from a bad local-minima and find a better one. We follow these inspiring ideas to address individual saturation and encourage SGD to explore more. The main differences are that we apply noise injection on CNN and impose noise on loss layer instead of previous layers. But different from adding noise on loss layer in DisturbLabel~\cite{Xie2016DisturbLabel}, a method that seems weird to disturb labels but indeed improves the performances of models, our work has a clear object of the delay of early softmax saturation by explicitly injecting noise into softmax activition. 

Another noise injection way is randomly transforming the input data, which is commonly referred to data augmentation, such as randomly cropping, flipping~\cite{Krizhevsky2012ImageNet, Wu2015A}, rotating~\cite{Laptev2016TI, Laptev2015Transformation} and jittering input data~\cite{Reed1992An, Reed1992Regularization}. And our work can also be interpreted as a way of data augmentation which will be discussed in the following discussion part.
\vspace{-0.5em}
\section{Early Individual Saturation}
\label{sec:individual}
\vspace{-0.5em}
In this section, we will give a toy example to describe the early individual saturation of softmax, which is always omitted, and analyse its impact on generalization. Define the $i$-th input data $x_{i}$ with the corresponding label $y_{i}$, $y_{i}\in[1\cdots C]$. Then processing training images with standard DCNN, we can obtain the cross-entropy loss and partial derivative as follows:
\vspace{-1em}
\begin{equation}\label{eq1}
L=-\frac{1}{N}\sum_{i}\log{P(y_{i}|x_{i})}=-\frac{1}{N}\sum_{i}\log{\frac{e^{f_{y_{i}}}}{\sum_{j}e^{f_{j}}}}
\vspace{-0.6em}
\end{equation}
\vspace{-0.6em}
\begin{equation}\label{eq2}
\frac{\partial{L}}{\partial{f_{j}}}=P(y_{i}=j|x_{i})-1\{y_{i}=j\}=\frac{e^{f_{j}}}{\sum_{k}e^{f_{k}}}-1\{y_{i}=j\}
\end{equation}
where $f_{j}$ refers to the j-th element of the softmax input vector $\mathbf{f}$, $j\in[1\cdots C]$, $N$ is the number of training images. $1\{condition\}=1$ if $condition$ is satisfied and $1\{condition\}=0$ if not.

\begin{figure}[!h]
\vspace{-0.6em}
  \centering
  \begin{minipage}{1\linewidth}
  \includegraphics[width=1\linewidth]{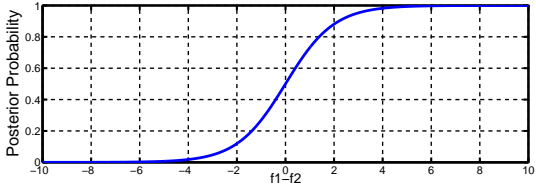}
  \vspace{-1em}
  \caption{Softmax activation function: $\frac{1}{1+e^{-(f_{1}-f_{2})}}$. X axis represents the difference between $f_{1}$ and $f_{2}$.}\label{fig:softmax}
  \end{minipage}
  \vspace{-1.2em}
\end{figure}

To simplify our analysis, we consider the problem of binary classification\footnote{Multi-classification complicates our analysis but has the same mechanism as binary scenario.}, where $y_{i}\in[1,2]$. Under binary scenario, we plot the softmax activation for class $1$ in Figure~\ref{fig:softmax}. Intuitively, the softmax activation is totally like sigmoid function. The standard softmax encourages $f_{1}>f_{2}$ in order to classify class $1$ correctly and can be regarded as a genius when its output $P(y_{i}=1|x_{i})=\frac{1}{1+e^{-(f_{1}-f_{2})}}$ is very close to $1$. In this case, the softmax output of data $x_{i}$ is saturated and we define this as individual saturation. Of course, making its softmax output close to $1$ is our ultimate goal of CNN training. However, we would like to achieve it at the end of SGD exploration not in the beginning or middle stage. Since, when optimizing CNN with gradients-based methods such as SGD, the prematurely saturated individual early stops contributing gradients to back-propagation due to negligible gradients, i.e. $P(y_{i}=1|x_{i})\approx 1, \frac{\partial{L}}{\partial{f_{y_{i}}}}\approx 0$ (see Eq.~\ref{eq2}). And with the saturated individuals number rising, the amount of contributing data decreases and, SGD has few chances to move around and is more likely to converge at a local minima, therefore, it is easy to be over-fitting and it requires extra data to recover. In short, the early saturated ones introduce short-lived gradients propagation which is not enough to help system converge at a 'global minima' (i.e. a better local-minima), so the early individual saturation is undesired.
\vspace{-1em}
\section{Noisy Softmax}
\label{sec:noisy softmax}
Based on the fact analysed in Section~\ref{sec:individual}, the short-lived gradients propagation caused by early individual saturation would not guide the robust learning. Thus the intuitive solution is to \textbf{set up 'barrier' along its way to satuation so as to postpone the early saturation behavior and produce rich and continuous gradients propagation}. Particularly, for training data point $(x_{i},y_{i})$, a simple way to achieve this is to factitiously reduce its softmax input $f_{y_{i}}$ (note that, it is theoretically same to enlarge $f_{j}, \forall{j}\neq y_{i}$, but it is so complex to operate). Moreover, many research works point out that adding noise gives the system chances to find 'global minima', such as injecting noise to sigmoid~\cite{Gulcehre2016Noisy}. We follow this inspiring idea to address the problem of early individual saturation. Therefore, our technique of slowing down the early saturation is to inject appropriate noise in softmax input $f_{y_{i}}$, and the resulting noise-associated one is as follows:
\vspace{-0.4em}
\begin{equation}\label{eq3}
  f_{y_{i}}^{noise}=f_{y_{i}}-n
\vspace{-0.6em}
\end{equation}
where $n=\mu + \sigma \xi$, $\xi \sim \mathcal{N}(0,1)$, $\mu$ and $\sigma$ are used to generate a wider family of noise from $\xi$. Intuitively, we would prefer $f_{y_{i}}^{noise}$ to be less than $f_{y_{i}}$ (because $f_{y_{i}}^{noise}>f_{y_{i}}$ will speed up saturation). Thus, we simply require noise $n$ to be always positive, and we have the following form:
\vspace{-0.3em}
\begin{equation}\label{eq4}
  f_{y_{i}}^{noise}=f_{y_{i}} - \sigma |\xi|
  \vspace{-0.3em}
\end{equation}
where noise $n$ has mean 0 and standard variance $\sigma$.

Moreover, we would like to make our noise annealed by controlling the parameter $\sigma$. Considering our initial thought, we intend to postpone the early saturation of $x_{i}$ instead of to not allow its saturation, implying that the initially larger noise is required to boost the exploration ability and later the relatively smaller noise is required for model convergence.

In standard Softmax layer (Figure~\ref{fig:decomposition}), $f_{y_{i}}$ is also the output of the fully connected component and can be written as $f_{y_{i}}=W^{T}_{y_{i}}X_{i}+b_{y_{i}}$ where $W_{y_{i}}$ is the $y_{i}$-th column of $W$, $X_{i}$ is the input feature of this layer from training data $x_{i}$ and $b_{y_{i}}$ is the basis. Since $b_{y_{i}}$ is a constant and $f_{y_{i}}$ mostly depends on $W^{T}_{y_{i}}X_{i}$, we construct our annealed noise by making $\sigma$ to be related to $W^{T}_{y_{i}}X_{i}$. In consideration of the fact that $W^{T}_{y_{i}}X_{i}=\|W_{y_{i}}\|\|X_{i}\|\cos{\theta_{y_{i}}}$(as shown in\cite{Liu2016Large}), where $\theta_{y_{i}}$ is the angle between vector $W_{y_{i}}$ and $X_{i}$, $\sigma$ should be a joint function of $\|W_{y_{i}}\|\|X_{i}\|$ and $\theta_{y_{i}}$ which hold amplitude and angular information respectively. Parameter $W_{y_{i}}$ followed by a loss function can be regarded as a linear classifier of class $y_{i}$. And this linear classifier uses cosine similarity to make angular decision boundary. As a result, with the converging of system, the angle $\theta_{y_{i}}$ between $W_{y_{i}}$ and $X_{i}$ will gradually decrease. Therefore, our annealed-noise-associated softmax input is formulated as:
\vspace{-0.2em}
\begin{equation}\label{eq5}
  f_{y_{i}}^{noise}=f_{y_{i}} - \alpha\|W_{y_{i}}\|\|X_{i}\|(1-\cos{\theta_{y_{i}}})|\xi|
  \vspace{-0.3em}
\end{equation}
where $\alpha\|W_{y_{i}}\|\|X_{i}\|(1-\cos{\theta_{y_{i}}})=\sigma$, and hyper-parameter $\alpha$ is used to adjust the scale of noise. In our annealed noise, we leverage $\|W_{y_{i}}\|\|X_{i}\|$ to make the magnitude of the noise and $f_{y_{i}}$ to be comparable, and use $(1-\cos{\theta_{y_{i}}})$ to adaptively anneal the noise. Notably, our early desaturation work implies that make softmax later saturated instead of non-saturated. We experimented with various function types of $\sigma$ and empirically found that this surprising simple formulation performs better. Putting Eq.~\ref{eq5} into original softmax, the Noisy Softmax loss is defined as:
\vspace{-0.3em}
\begin{equation}\label{eq6}
  L=-\frac{1}{N}\sum_{i}\log{\frac{e^{f_{y_{i}} - \alpha\|W_{y_{i}}\|\|X_{i}\|(1-\cos{\theta_{y_{i}}})|\xi|}}{\sum_{j\neq{y_{i}}}e^{f_{j}}+e^{f_{y_{i}}-\alpha\|W_{y_{i}}\|\|X_{i}\|(1-\cos{\theta_{y_{i}}})|\xi|}}}
\vspace{-0.8em}
\end{equation}
\begin{figure*}[t]
  \vspace{-1em}
  \centering
  \begin{minipage}[t]{0.48\linewidth}
  \centering
  \includegraphics[width=1\linewidth]{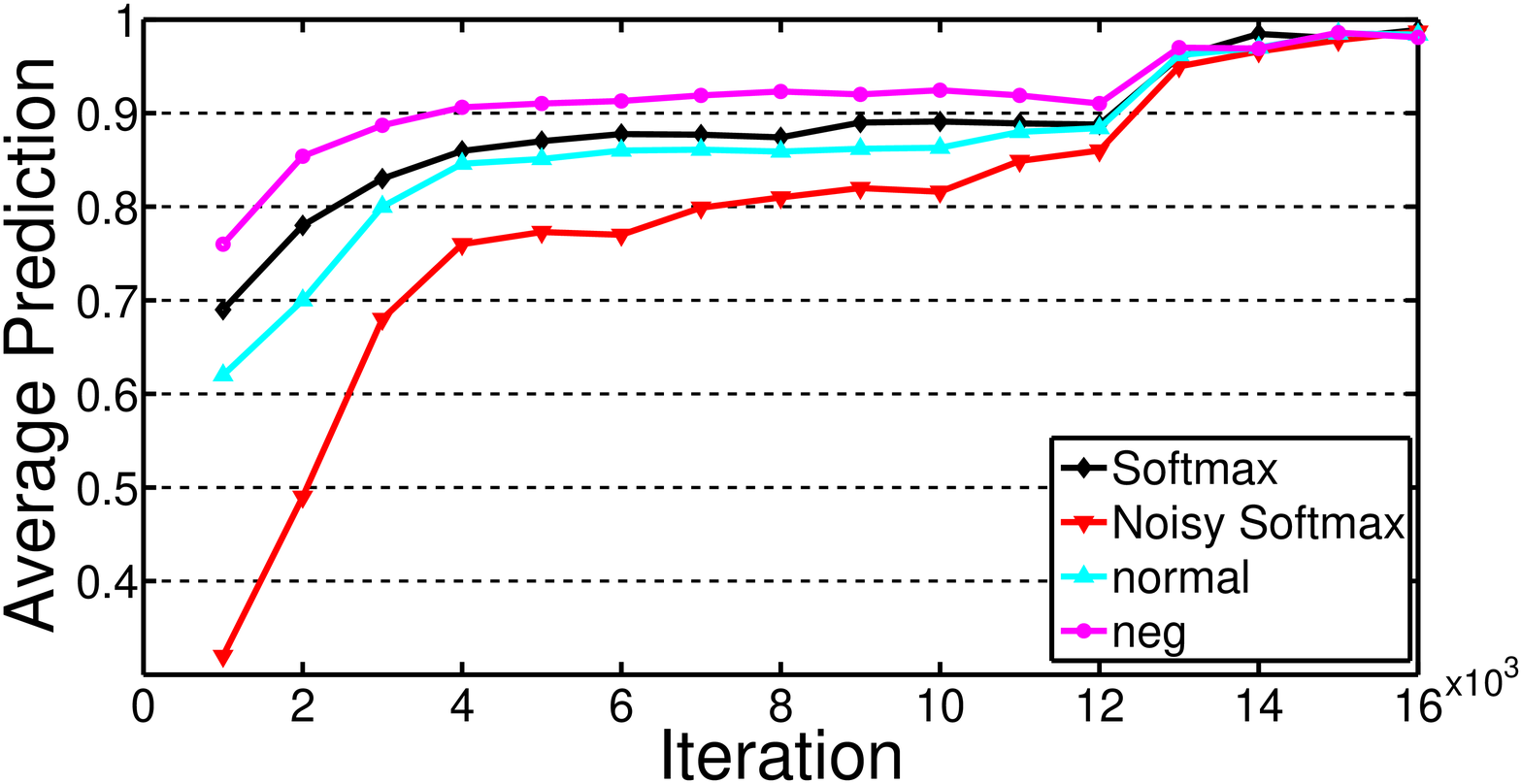}\\
  \caption{\label{fig3}Saturation status vs. iteration with different formulations of noise. Normal and Neg represent $normal$ noise and $negative$ noise respectively. $\alpha^{2}$ is set to 0.1 in our experiments.}
  \end{minipage}
  \begin{minipage}[t]{0.48\linewidth}
  \centering
  \includegraphics[width=1\linewidth]{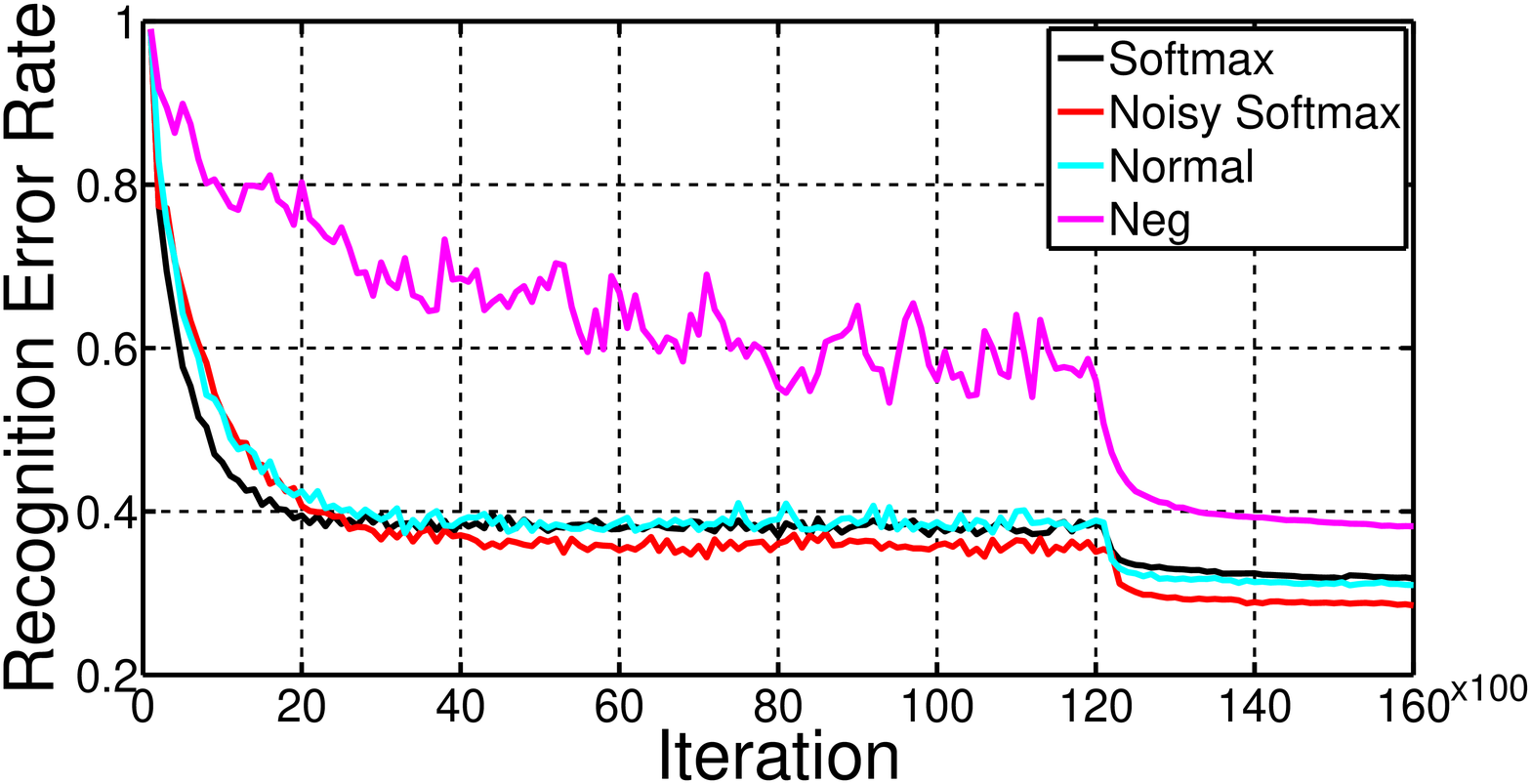}\\
  \caption{\label{fig6}CIFAR100 testing error vs. iteration with different formulations of noise. Normal and Neg represent $normal$ noise and $negative$ noise respectively. $\alpha^{2}$ is set to 0.1 in our experiments.}
  \end{minipage}
  \vspace{-1.2em}
\end{figure*}

\textbf{Optimization}. We use Eq.~\ref{eq6} throughout our experiments and optimize our model with the commonly used SGD. Thus we need to compute the forward and backward propagation, and $\cos{\theta_{y_{i}}}$ is required to be replaced with $\frac{W^{T}_{y_{i}}X_{i}}{\|W_{y_{i}}\|\|X_{i}\|}$. For forward and backward propagation, the only difference between Noisy Softmax loss and standard softmax loss exists in $f_{y_{i}}$. For example, in forward propagation, $\forall{j}\neq{y_{i}}$ $f_{j}$ is computed as the same as original softmax while $f_{y_{i}}$ is replaced with $f_{y_{i}}^{noise}$. In backward propagation, $\frac{\partial{L}}{\partial{X_{i}}}=\sum_{j}\frac{\partial{L}}{\partial{f_{j}}}\frac{\partial{f_{j}}}{\partial{X_{i}}}$ and $\frac{\partial{L}}{\partial{W_{y_{i}}}}=\sum_{j}\frac{\partial{L}}{\partial{f_{j}}}\frac{\partial{f_{j}}}{\partial{W_{y_{i}}}}$, only when $j=y_{i}$ the computations of $\frac{\partial{f_{j}}}{\partial{X_{i}}}$ and $\frac{\partial{f_{j}}}{\partial{W_{y_{i}}}}$ are not the same as original softmax which are listed as follows:
\begin{align}
\vspace{-0.8em}
  \frac{\partial{f_{y_{i}}^{noise}}}{\partial{X_{i}}}&=W_{y_{i}}-\alpha|\xi|(\frac{X_{i}\|W_{y_{i}}\|}{\|X_{i}\|}-W_{y_{i}})\label{eq7}\\
  \frac{\partial{f_{y_{i}}^{noise}}}{\partial{W_{y_{i}}}}&=X_{i}-\alpha|\xi|(\frac{W_{y_{i}}\|X_{i}\|}{\|W_{y_{i}}\|}-X_{i})\label{eq8}
  \vspace{-0.8em}
\end{align}
For simplicity, we leave out $\frac{\partial{L}}{\partial{f_{j}}}$ and $\frac{\partial{f_{j}}}{\partial{X_{i}}}$, $\frac{\partial{f_{j}}}{\partial{W_{y_{i}}}} (\forall{j}\neq{y_{i}})$ since they are the same for both Noisy Softmax and original softmax. In short, except for when $j=y_{i}$, the overall computation of Noisy Softmax is similar with original softmax.
\vspace{-1em}
\section{Discussion}
\subsection{The Effect of Noise Scale $\bm{\alpha}$}
In Noisy Softmax, the scale of annealed noise is largely determined by the hyper-parameter $\alpha$. Here we can imagine that, when $\alpha = 0$ in the $0$ noise limit, Noisy Softmax is the same with ordinary softmax. Then individual saturation will turn up and SGD solver has a high chance to converge at a local-minima. Without extra data for training, the model will be easily over-fitting. However, when $\alpha$ is large enough, large gradients are obtained since backpropagating through $f_{y_{i}}^{noise}$ gives rise to large derivatives. So the algorithm just see the noise instead of real signal and move around anywhere blindly. Hence, a relatively small $\alpha$ is required to aid the generalization of model.

\begin{figure*}[t]
  \vspace{-1em}
  \centering
  \begin{minipage}{0.48\linewidth}
  \centering
  \includegraphics[width=1\linewidth]{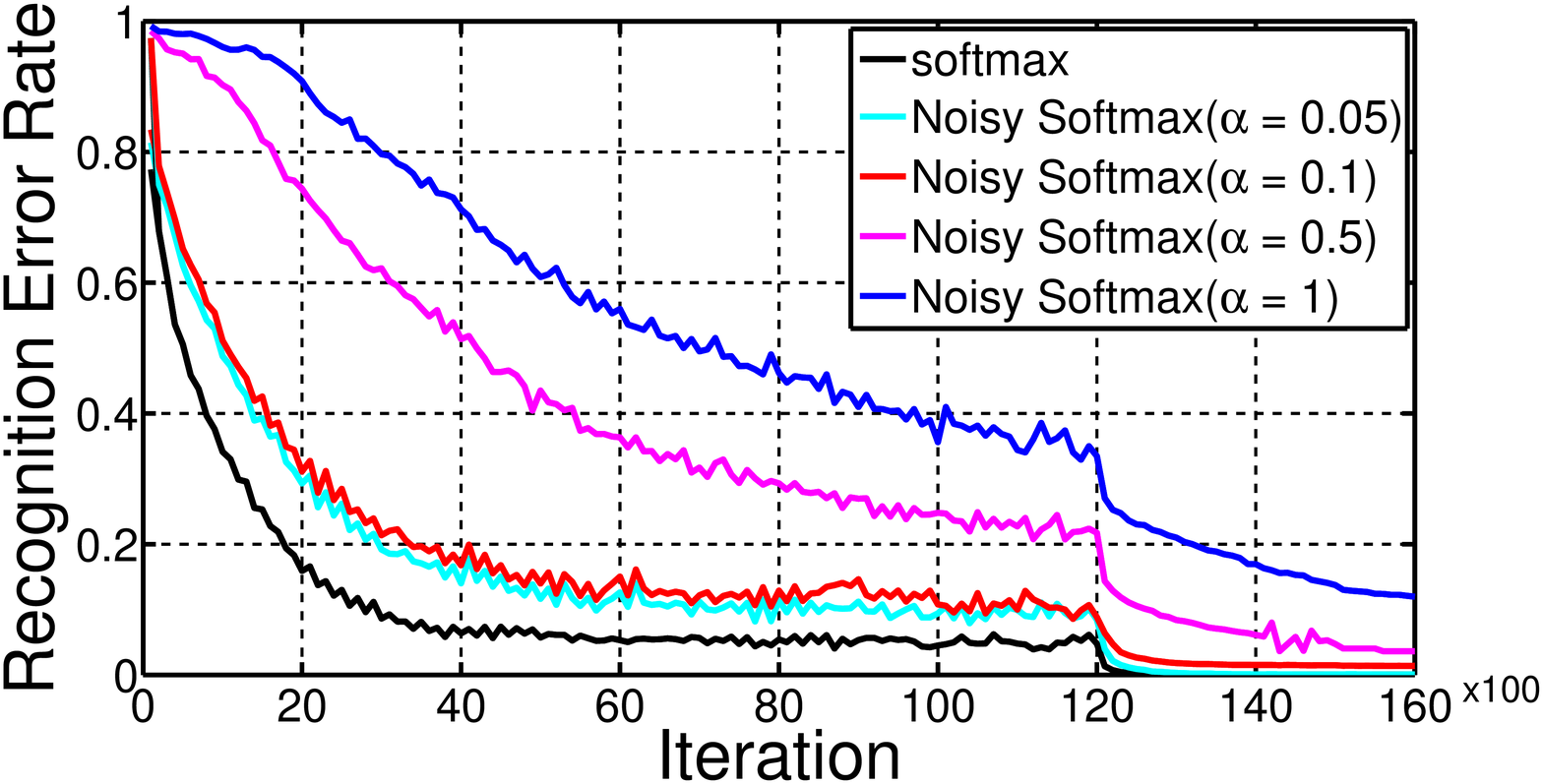}\\
  \vspace{-0.6em}
  \caption{CIFAR100 training error vs. iteration with different $\alpha$.}\label{fig4}
  \end{minipage}
  \begin{minipage}{0.48\linewidth}
  \centering
  \includegraphics[width=1\linewidth]{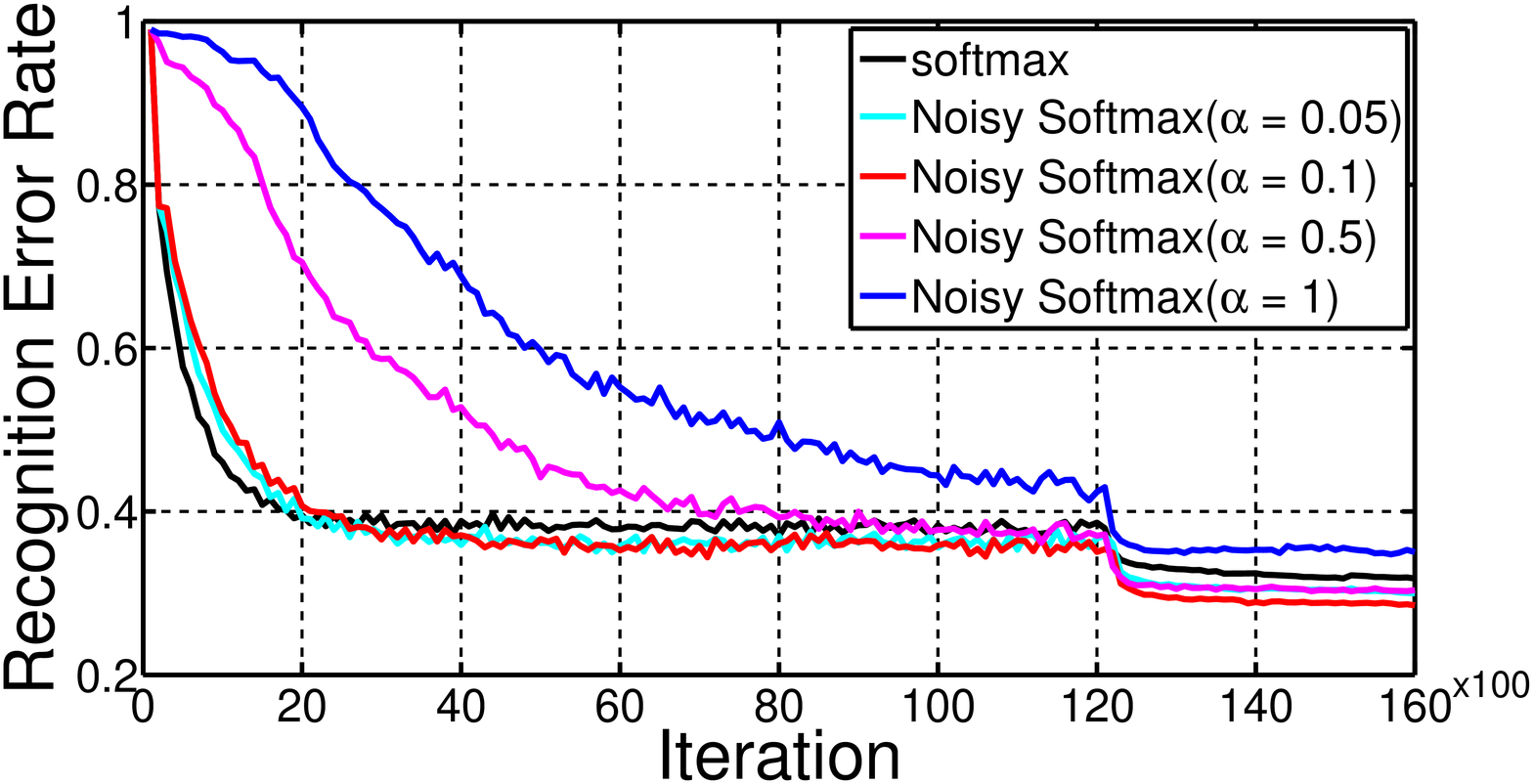}\\
  \vspace{-0.6em}
  \caption{CIFAR100 testing error vs. iteration with different $\alpha$.}\label{fig5}
  \end{minipage}
\vspace{-1.2em}
\end{figure*}

We evaluate the performances of Noisy Softmax with different $\alpha$ on several datasets. Note that, the value of $\alpha$ is not carefully adjusted and we make $\alpha=0$ (i.e. softmax) as our baseline. And these comparison results are listed in Table~\ref{tab2},~\ref{tab3}and~\ref{tab4}. One can observe that Noisy Softmax with a relatively appropriate $\alpha$ (e.g. $\alpha^{2}=0.1$) obtains better recognition accuracy than ordinary softmax on all of the datasets. To be intuitional, we summarize the results on CIFAR100 in Figure~\ref{fig5}. When $\alpha^{2}=0.1$, our method outperforms the original softmax. This demonstrates that our Noisy Softmax does improve the generalization ability of CNN by encouraging the SGD solver to be more exploratory and to converge at a 'global-minima'. When $\alpha$ rises to $1$, the large noise causes the network to converge slower and produces worse performance than baseline as well. Since the large noise drowns the helpful signal and solver just sees the noise.
\subsection{Saturation Study}\label{set:saturation}
To illuminate the significance of early softmax desaturation based on non-negative noise injection, we investigate the impacts of different formulations of noise, such as $normal$ noise $n=\sigma\xi$ and $negative$ noise $n=-\sigma|\xi|$ ($\sigma$ is the same as in Eq.~\ref{eq5}), on individual saturation. From the formulations of noise, we can imagine that there will be more saturated instances when training with $negative$ noise (which is a counterexample). To intuitively analyse the saturation state, we compute the average possibility prediction over the entire training set as follows:
\vspace{-0.7em}
\begin{equation}\label{eq9}
  \overline{P}=\frac{1}{N}\sum_{j=1}^{C}\sum_{i=1}^{N_{j}}P(y_{i}|x_{i})
  \vspace{-0.7em}
\end{equation}
where $C$ is the number of classes and $N_{j}$ is the number of images within the $j$-th class. 
Figure~\ref{fig3} and~\ref{fig6} show the saturation status of different noise and testing error rates on CIFAR100 respectively.

From the results in Figure~\ref{fig3}, one can observe that when training with original softmax or $negative$ noise, the average prediction rises quickly to a relatively high level, almost $0.9$, implying that the early individual saturation is serious, and finally goes up to nearly $1$. Moreover, the average prediction of $negative$ noise is higher than that of softmax, implying that the early individual saturation is deteriorated since many instances are factitiously mapped to saturated ones. From the results in Figure~\ref{fig6}, one can observe that the testing error of $negative$ noise drops slowly and finally achieves a relatively high level, nearly 37\%, verifying that the exploration of SGD is seriously impeded by early individual saturation. In $normal$ noise case, the testing error and the trend of average prediction rise are similar with original softmax, shown in Figure~\ref{fig6} and~\ref{fig3} respectively, since the expectation $E(n)$ is close to zero.

In contrast, when training with Noisy Softmax, the average prediction rises slowly and is much lower than original softmax at the early training stage, shown in Figure~\ref{fig3}, verifying that the early individual saturation behavior is significantly avoided. And from the results in Figure~\ref{fig6}, one can observe that Noisy Softmax outperforms the baseline and significantly improves the performance to 28.48\% testing error rate. Note that, after 3,000 iterations, our method achieves better testing error result but lower average-prediction, demonstrating that the early desaturation of softmax gives SGD solver chances to traverse more portions of parameter space for optimal solution. As the noise level is decreased, it will prefer a better local-minima where signal gives strong response to SGD. Then the solver will spend more time to explore this region and converge, which can be regarded as 'global-minima', in a finite number of steps.

In summary, injecting non-negative noise $n=\sigma|\xi|$ in softmax does prevent the early individual saturation and further improve the generalization ability of CNN when optimized by standard SGD.
\subsection{Annealed Noise Study}
\begin{table}[b]
\vspace{-1.5em}
  \centering
    \begin{tabular}{|c|c|c|c|}
    \hline
    $\alpha^{2}$    & Noisy Softmax & $free$ & $amplitude$\\
    \hline
    0     & 31.77 & 31.77 & 31.77 \\
    \hline
    0.05  & 29.99 & 31.43 & 30.96 \\
    \hline
    0.1   & \bf{28.48} & 31.04 & 29.97 \\
    \hline
    0.5   & 30.22 & 30.88 & fail \\
    \hline
    1     & 35.23 & 31.20 & fail \\
    \hline
    \end{tabular}%
  \caption{Testing error rates(\%) vs. different noise on CIFAR100.}
  \label{tab5}%
\end{table}%
When addressing the early individual saturation, the key idea is to add annealed noise. In order to highlight the superiority of our annealed noise described in Sec.~\ref{sec:noisy softmax}, we compare it to $free$ noise $n=\alpha|\xi|$ and $amplitude$ noise $n=\alpha\|W_{y_{i}}\|\|X_{i}\||\xi|$. We evaluate them on CIFAR100 and the results are listed in Table~\ref{tab5}. From the results, it can be observed that our Noisy Softmax outperforms the other two formulations of noise. In $free$ noise case, where $\sigma$ (described in Sec.~\ref{sec:noisy softmax}) is set to a fixed value $\alpha$ and the noise is totally independent, although adding this noise is a desaturation operation the accuracy gain over the baseline is small, since it desaturates softmax worse not according to the magnitude of softmax input, in another word it cannot suit the remedy to the case. In $amplitude$ noise case, where $\sigma$ is set to $\alpha\|W_{y_{i}}\|\|X_{i}\|$, the subtractive noise is prudent due to considering the level of softmax input, thus yielding better accuracy gain than $free$ noise. While it still is worse than Noisy Softmax. Because, in both Noisy Softmax and $amplitude$ noise cases, as the exploration going, the 'globally better' region of parameter space has been seen by SGD and it is time to be patient in exploring this region, in another word smaller noise is required. Noisy Softmax holds this idea by annealing the noise but, in $amplitude$ noise case, the level-unchanging noise seems a little large at this time and further causes difficulty in detailed learning. Reviewing the formulation of our annealed noise, one can observe that our annealed noise is constructed by combining $\theta_{y_{i}}$, a time identifier, into $amplitude$ noise. With time function $1-\cos{\theta_{y_{i}}}$ injected, the noise will be adaptively decreased.
\subsection{Regularization Ability}
\label{sec:prevent}
We find experimentally that Noisy Softmax can regularize the CNN model by preventing over-fitting. Figure~\ref{fig4} and~\ref{fig5} show the recognition accuracy results on CIFAR100 dataset with different $\alpha$. One can observe that without noise injection (i.e. $\alpha=0$), the training recognition error drops fast to quite a low level, almost $0\%$, however the testing recognition error stops falling at a relatively high level, nearly $31.77\%$. Conversely, when $\alpha^{2}$ is set to an appropriate value such as 0.1, the training error drops slower and is much higher than the baseline. But the testing error reaches a lower level, nearly $28.48\%$, and still has a trend of decreasing. Even when $\alpha^{2}=0.5$, the training error is higher but the testing error becomes lower as well, nearly $30.22\%$. This demonstrates that encouraging SGD to converge at a better local-minima indeed prevent over-fitting and Noisy Softmax has a strong regularization ability. 

\begin{figure}[t]
  \vspace{-.2em}
  \centering
  \begin{minipage}{0.9\linewidth}
  \centering
  \includegraphics[width=1\linewidth]{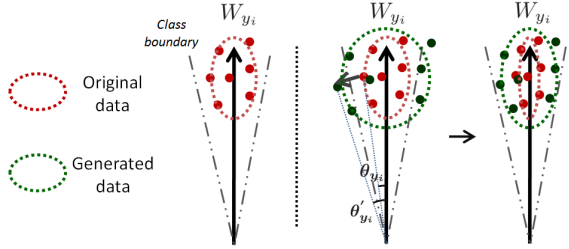}\\
  \caption{Geometric Interpretation of data augmentation.}\label{fig7}
  \end{minipage}
  \vspace{-1.2em}
\end{figure}

As analysed above, our Noisy Softmax can be regarded as a kind of regularization technique for preventing over-fitting by making SGD to be more exploratory. Here we will analyse this regularization ability from another data augmentation perspective, which has a profound physical interpretation. Under the original case, the softmax input coming from data point $(x_{i},y_{i})$ is $f_{y_{i}}=\|W_{y_{i}}\|\|X_{i}\|\cos{\theta_{y_{i}}}$ (we omit the constant $b_{y_{i}}$ for simplicity). Now we consider a new input $(x_{i}^{'},y_{i})$, where $\|X_{i}^{'}\|=\|X_{i}\|$ and the angle $\theta_{y_{i}}^{'}$ between vector $W_{y_{i}}$ and $X_{i}^{'}$ is $\arccos((1+\alpha|\xi|)\cos{\theta_{y_{i}}}-\alpha|\xi|)$. Thus we have $f_{y_{i}}^{'}=\|W_{y_{i}}\|\|X_{i}^{'}\|\cos{\theta_{y_{i}}^{'}}=W_{y_{i}}^{T}X_{i}-\alpha\|W_{y_{i}}\|\|X_{i}\|(1-\cos{\theta_{y_{i}}})|\xi|=f_{y_{i}}^{noise}$, implying that $f_{y_{i}}^{noise}$ can be regarded as coming from a new data point $(x_{i}^{'},y_{i})$. Notably, since $\theta_{y_{i}}^{'}>\theta_{y_{i}}$, these generated data have many boundary examples which are much helpful for discriminative feature learning, illustrated in Figure~\ref{fig7}. In summary, generating the noisy input $f_{y_{i}}^{noise}$ is equivalent to generating new training data, which is an efficient way of data augmentation.

To verify our discussion above, we evaluate Noisy Softmax on two subsets of the MNIST dataset, which have only $600(1\%)$ and $6000(10\%)$ training instances respectively. Our CNN configuration is shown in Table~\ref{tab1}. With the same training strategy in Section~\ref{sec:training}, we achieve $3.82\%$ and $1.30\%$ testing error rates on the original testing set, respectively. Meanwhile, in both cases, the training error rates quickly drop to nearly $0\%$ which show that the over-fitting turns up. However, when training with Noisy Softmax $(\alpha^{2}=0.5)$, we obtain $2.46\%$ and $0.93\%$ testing error rates, respectively. This demonstrates that Noisy Softmax improves the generalization ability of CNN with implicit data augmentation. And from the accuracy improvements on these two subsets and CIFAR100 (which has 500 instances per class), it acts as an effective algorithm especially in the case that the amount of training data is limited.
\subsection{Relationship to Other Methods}
\textbf{Multi-task learning}: combining several tasks to one system does improve the generalization ability~\cite{Seltzer2013Multi}. Considering a multi-task learning system with input data $x_{i}$, the overall object loss function is a combination of several sub-object loss functions, written as $L=\sum_{j}L_{j}(\vartheta_{0},\vartheta_{j},x_{i})$, where $\vartheta_{0}$ and $\vartheta_{j},j\in[1,2,\cdots]$ are generic parameters and task-specific parameters respectively. Optimizing with standard SGD, generic parameters $\vartheta_{0}$ are updated as $\vartheta_{0}=\vartheta_{0}-\gamma(\sum_{j}\frac{\partial{L_{j}}}{\partial{\vartheta_{0}}})$, $\gamma$ is the learning rate. While in Noisy Softmax, from an overall training perspective, our loss function can also be regarded as a combination of many noise-dependent changing losses $L_{noise_{k}}=-\log{\frac{e^{f_{y_{i}}}}{\sum_{c\neq{y_{i}}}e^{f_{c}}+e^{f_{y_{i}}^{noise}}}}+\alpha|\xi|_{k}\|W_{y_{i}}\|\|X_{i}\|(1-\cos{\theta_{y_{i}}}),k\in[1,m]$, i.e. $L=\sum_{k=1}^{m}L_{noise_{k}}$ where $m$ is an uncertain number and is related to noise scale and iteration number. Thus, the overall contribution to system can be regarded as $\vartheta=\vartheta-\gamma(\sum_{k=1}^{m}\frac{\partial{L_{noise_{k}}}}{\partial\vartheta})$. So our method can be regarded as a special case of multi-task learning, where the task-specific parameters are shared across tasks.

However, in multi-task learning system, factitiously designing task-specific losses is prohibitively expensive and the number of tasks is limited and small. While in Noisy Softmax training procedure, the model is constrained by many randomly generated tasks (quantified by $L_{noise_{k}}$). Thus, training a model with Noisy Softmax can be regarded as training with massive tasks that are very costly and often infeasible to design in original multi-task learning system.


\textbf{Noise injection}: some research works inject noise in previous layers of neural networks such as in neuron activation functions ReLU~\cite{by2015Rectified,Bengio2013Estimating} and sigmoid~\cite{Gulcehre2016Noisy}, in weights~\cite{Graves2011Practical,Blundell2015Weight} and gradients~\cite{Neelakantan2015Adding}. We emphasize that Noisy Softmax adds noise on a single loss layer instead of on many previous layers, which is more convenient and efficient for implementation and model training, and is applied in DCNNs. Intrinsically different from DisturbLabel~\cite{Xie2016DisturbLabel}, where noise is produced by disturbing labels and also exerts effect on loss layer, Noisy Softmax starts from a clear object of early softmax desaturation and the noise is adaptively annealed and injected in a explicit manner.

\textbf{Desaturation}: 
many other desaturation works, such as replacing sigmoid with ReLU~\cite{Glorot2010Deep} and building skip connections between layers~\cite{Srivastava2015Highway, He2015Deep, He2016Identity, Huang2016Densely}, solve the problems of gradients vanishing which happen in bottom layers. While our Noisy Softmax solves the problem of early gradients vanishing in the top layer (i.e. loss layer) which is caused by the early individual saturation. We emphasize that solving the early gradients vanishing in top layer is crucial to parameters update and model optimization, since top layer is the source of gradients propagation. In summary, by postponing the early individual saturation we can obtain continuous gradients propagation from the top layer and further encourage SGD to be more exploratory.
\begin{table*}[t]
  \centering
    \begin{tabular}{|c|c|c|c|c|c|}
    \hline
    Layer & MNIST(for Sec.~\ref{sec:prevent}) & MNIST & CIFAR10/10+ & CIFAR100 & LFW/FGLFW/YTF \\
    \hline
    \hline
    Block1 &[3x3,40]x2 & [3x3,64]x3 & [3x3,64]x4 & [3x3,96]x4 & [3x3,64]x1 \\
    \hline
    Pool1 & \multicolumn{5}{c|}{Max [2x2], stride 2} \\
    \hline
    Block2 & [3x3,60]x1 & [3x3,64]x3 & [3x3,128]x4 & [3x3,192]x4 & [3x3,128]x1 \\
    \hline
    Pool2 & \multicolumn{5}{c|}{Max [2x2], stride 2} \\
    \hline
    Block3 & [3x3,60]x1 & [3x3,64]x3 & [3x3,256]x4 & [3x3,384]x4 & [3x3,256]x2 \\
    \hline
    Pool3 & \multicolumn{5}{c|}{Max [2x2], stride 2} \\
    \hline
    Block4 & - & -     & -     & -     & [3x3,512]x3,padding 0 \\
    \hline
    Fully Connected & 100 & 256   & 512   & 512   & 3000 \\
    \hline
    \end{tabular}%
    \caption{CNN architectures for different benchmark datasets. Blockx denotes a container of several convolution components with the same configuration. E.g. [3x3, 64]x4 denotes 4 cascaded convolution layers with 64 filters of size 3x3.}
  \label{tab1}%
\vspace{-1.5em}
\end{table*}
\vspace{-1em}
\section{Experiments and Results}
\label{sec:experiments}
\vspace{-0.5em}
We evaluate our proposed Noisy Softmax algorithm on several benchmark datasets, including MNIST~\cite{LecunThe}, CIFAR10/100~\cite{Krizhevsky2012Learning},
LFW~\cite{Huang2008Labeled}, FGLFW~\cite{Nanhai2016Fine} and YTF~\cite{Wolf2011Face}. Note that, in all of our experiments, we only use a single model for the evaluation of Noisy Softmax, and both softmax and Noisy Softmax in our experiments use the same CNN architecture shown in Table~\ref{tab1}.
\vspace{-0.6em}
\subsection{Architecture Settings and Implementation}
\label{sec:training}
As VGG~\cite{Simonyan2015Very} becomes a commonly used CNN architecture, the cascaded layers with small size filters gradually take the place of single layer with large size filters. Since these cascaded layers have less parameters, lower computational complexity and stronger representation ability compared to the single layer. E.g. a single 5x5 convolution layer is replaced with 2 cascaded 3x3 convolution layers. Inspired by \cite{Simonyan2015Very,Liu2016Large}, we design the architectures as shown in Table~\ref{tab1}. In convolution layers, both the stride and padding are set to 1 if not specified. In pooling layers, we use 2x2 max-pooling filter with stride 2. We adopt the piece-wise linear functions PReLU~\cite{He2015Delving} as our neuron activation functions. Then we use the weight initialization~\cite{He2015Delving} and batch normalization~\cite{Ioffe2015Batch} in our networks. All of our experiments are implemented by Caffe library~\cite{Turchenko2014Caffe} with our own modifications. We use standard SGD to optimize our CNNs and the batch sizes are 256 and 200 for object experiments and face experiments, respectively. For data preprocessing, we just perform the mean substraction.

\textbf{Training}. In object recognition tasks, the initial learning rate is 0.1 and is divided by 10 at 12k. The total iteration is 16k. Note that, although we train our CNNs with coarsely adjusted learning rate, the results of all experiments are impressive and consistent, verifying the effectiveness of our method. For face recognition tasks, we start with a learning rate of 0.01, divide it by 5 when the training loss does not drop.

\textbf{Testing}. We use the original softmax to classify the testing data in object datasets. In face datasets, we evaluate it with the cosine distance rule after PCA reduction for face recognition.
\vspace{-0.6em}
\subsection{Evaluation on MNIST Dataset}
\textbf{MNIST}~\cite{LecunThe} contains 60,000 training samples and 10,000 testing samples. These samples are uniformly distributed over 10 classes. And all samples are 28x28 gray images. Our CNN network architecture is shown in Table~\ref{tab1} and we use a 0.001 weight decay. The results of the state-of-the-art methods and our proposed Noisy Softmax with different $\alpha$ are listed in Table~\ref{tab2}.

From the results, our Noisy Softmax ($\alpha^{2} = 0.1$) not only outperforms the original softmax over the same architecture, but also achieves the competitive performance compared to the state-of-the-art methods. It can also be observed that Noisy Softmax produces consistent accuracy gain with coarsely adjusted $\alpha^{2}$, such as 0.05, 0.1 and 0.5, and our method achieves the same accuracy with DisturbLabel~\cite{Xie2016DisturbLabel} which adds dropout to several layers, demonstrating the effectiveness of our technique.
\subsection{Evaluation on CIFAR Datasets}\label{sec:cifar}
\textbf{CIFAR}~\cite{Krizhevsky2012Learning} has two evaluation protocols over 10 and 100 classes respectively. CIFAR10/100 has 50,000 training samples and 10,000 testing samples, all the samples are 32x32 RGB images. And these images are uniformly distributed over 10 or 100 classes. We use different CNN architectures in CIFAR10 and CIFAR100 experiments, and these network configurations are shown in Table~\ref{tab1}.

We evaluate our method on CIFAR10 and CIFAR100, as these results are shown in Table~\ref{tab3}. For data augmentation, we perform a simple method: randomly crop a 30*30 image. From our experimental results, one can observe that Noisy Softmax ($\alpha^{2}=0.1$) outperforms all of the other methods on these two datasets. And it improves nearly 1\% and more than 3\% accuracies over the baseline on CIFAR10 and CIFAR100 respectively. 
\subsection{Evaluation on Face Datasets}
\begin{table*}[t]
  \begin{minipage}{.35\linewidth}
  \centering
    \begin{tabular}{|c|c|}
    \hline
    Method & MNIST \\
    \hline
    \hline
    CNN~\cite{Jarrett2009What}   & 0.53 \\
    NiN~\cite{Lin2013Network}   & 0.47 \\
    Maxout~\cite{Goodfellow2013Maxout} & 0.45 \\
    DSN~\cite{lee2015deeply}   & 0.39 \\
    R-CNN~\cite{liang2015recurrent} & 0.31\\
    GenPool~\cite{Lee2015Generalizing} & 0.31 \\
    DisturbLabel~\cite{Xie2016DisturbLabel} & 0.33 \\
    \hline
    \hline
    Softmax & 0.43 \\
    Noisy Softmax ($\alpha^{2}=1$) & 0.42 \\
    Noisy Softmax ($\alpha^{2}=0.5$) & \bf{0.33} \\
    Noisy Softmax ($\alpha^{2}=0.1$) & \bf{0.33} \\
    Noisy Softmax ($\alpha^{2}=0.05$) & 0.37 \\
    \hline
    \end{tabular}%
    \caption{Recognition error rates (\%) on MNIST.}
  \label{tab2}%
  \end{minipage}
  \begin{minipage}{.65\linewidth}
  \centering
    \begin{tabular}{|c|c|c|c|}
    \hline
    Method & CIFAR10 & CIFAR10+ & CIFAR100 \\
    \hline
    \hline
    NiN~\cite{Lin2013Network}   & 10.47 & 8.81 & 35.68 \\
    Maxout~\cite{Goodfellow2013Maxout} & 11.68 & 9.38 & 38.57 \\
    DSN~\cite{lee2015deeply}   & 9.69 & 7.97 & 34.57 \\
    All-CNN~\cite{Springenberg2014Striving} & 9.08 & 7.25  & 33.71 \\
    R-CNN~\cite{liang2015recurrent} & 8.69 & 7.09 & 31.75 \\
    ResNet~\cite{He2016Identity} & N/A & 6.43 & N/A \\
    DisturbLabel~\cite{Xie2016DisturbLabel} & 9.45 & 6.98 & 32.99 \\
    \hline
    \hline
    Softmax & 8.11 & 6.98 & 31.77 \\
    Noisy Softmax ($\alpha^{2}=1$) & 9.09 & 8.77 & 35.23 \\
    Noisy Softmax ($\alpha^{2}=0.5$) & 7.84 & 7.13 & 30.22 \\
    Noisy Softmax ($\alpha^{2}=0.1$) & \bf{7.39} & \bf{6.36} & \bf{28.48} \\
    Noisy Softmax ($\alpha^{2}=0.05$) & 7.58 & 6.61 & 29.99 \\
    \hline
    \end{tabular}%

    \caption{Recognition error rates (\%) on CIFAR datasets. + denotes data augmentation.}
  \label{tab3}%
  \end{minipage}
\vspace{-0.6em}
\end{table*}%
\begin{table*}[t]
  \centering
    \begin{tabular}{|c|c|c|ccc|c|c|}
    \hline
    Method & Images & Models & LFW   & Rank-1 & DIR@FAR=1\% & FGLFW & YTF\\
    \hline
    \hline
    FaceNet~\cite{Schroff2015FaceNet} & 200M*  & 1     & 99.65 & -     & -     & - & 95.18\\
    DeepID2+~\cite{Sun2014Deeply} & 300k*  & 1     & 98.7  & -     & -     & - & 91.90\\
    DeepID2+~\cite{Sun2014Deeply} & 300k*  & 25    & 99.47 & 95.00    & 80.70  & - & 93.20\\
    Sparse~\cite{Sun2015Sparsifying} & 300k*  & 1     & 99.30  & -     & -     & - & 92.70\\
    VGG~\cite{Parkhi2015Deep}   & 2.6M  & 1     & 97.27 & 74.10  & 52.01 & 88.13 & 92.80\\
    \hline
    \hline
    WebFace~\cite{Yi2014Learning} & WebFace & 1     & 97.73 & -     & -     & - & 90.60\\
    Robust FR~\cite{Ding2015Robust} & WebFace & 1     & 98.43 & -     & -     & - & -\\
    Lightened CNN~\cite{Wu2015A} & WebFace & 1     & 98.13 & 89.21 & 69.46 & 91.22 & 91.60\\
    \hline
    \hline
    Softmax & WebFace$^{+}$ & 1     & 98.83 & 91.68 & 69.51 & 92.95 & 94.22\\
    Noisy Softmax($\alpha^{2}=0.1$) & WebFace$^{+}$ & 1 & \bf{99.18}  & \bf{92.68} & \bf{78.43} & \bf{94.50} & \bf{94.88}\\
    Noisy Softmax($\alpha^{2}=0.05$) & WebFace$^{+}$ & 1 & 99.02&92.24& 75.67 &94.02& 94.51\\
    \hline
    \end{tabular}%
    \caption{Recognition accuracies (\%) on LFW, FGLFW and YTF datasets. * denotes the images are not publicly available and $^{+}$ denotes data expansion. In LFW, closed-set and open-set accuracies are evaluated by Rank-1 and DIR@FAR=1 respectively.}
  \label{tab4}%
\vspace{-1.3em}
\end{table*}%
\textbf{LFW}\cite{Huang2008Labeled} contains 13,233 images from 5749 celebrities. Under unrestricted conditions, it provides 6,000 face pairs for verification protocol and, closed-set and open-set for identification protocol adopted in~\cite{best2014unconstrained}.

\textbf{FGLFW}~\cite{Nanhai2016Fine} is a derivative of LFW, implying that the images are all coming from LFW but the face pairs are difficult to classify whether they are from the same person. It is a light and sweet dataset for performance evaluation due to the simple verification protocol but challenging face pairs.

\textbf{YTF}~\cite{Wolf2011Face} provides $5,000$ video pairs for face verification. We use the average representation of 100 randomly selected samples from each video for evaluation.

For data preprocessing, we align and crop images based on eyes and mouth centers, yielding $104\times96$ RGB images. Our CNN configuration is shown in Table~\ref{tab1}, here we add element-wise maxout layer~\cite{Wu2015A} after the 3,000-dimensional fully connected layer, yielding a 1,500-dimensional output, and contrastive loss is applied on this output as in DeepID2~\cite{Sun2014Deep}. Then we train a single CNN model with outside data from the publicly available CASIA-WebFace dataset~\cite{Yi2014Learning} and our own collected data (about 400k from 14k identities). Extract the features for each image and its horizontally flipped one, then compute a mean feature vector as the representation. From the results shown in Table~\ref{tab4}, one can observe that Noisy Softmax ($\alpha^{2}=0.1$) improves the performance over the baseline, and the result is also comparable to the current state-of-the-art methods with private data and even model ensemble. In addition, we further improve our results to $\bm{99.31\%,94.43\%,82.50\%,94.88\%,95.37\%}$ (listed in the same protocol order as in Table.~\ref{tab4}) by two models ensemble.
\section{Conclusion}
In this paper, we propose Noisy Softmax to address the early individual saturation by injecting annealed noise to the softmax input. It is a way of early softmax desaturation by postponing the early individual saturation. We show that our method can be easily performed as a drop-in replacement for standard softmax and is easier to optimize. It significantly improves the performances of CNN models, since the early desaturation operation indeed exerts much effect on parameter update during back-propagation and furthermore improves the generalization ability of DCNNs. Empirical studies verify the superiority of softmax desaturation. Meanwhile, it achieves state-of-the-art or competitive results on several datasets.
\vspace{-1em}
\section{Acknowledgments}
This work was partially supported by the National Natural Science Foundation of China (Project 61573068, 61471048, 61375031 and 61532006), Beijing Nova Program under Grant No. Z161100004916088, the Fundamental Research Funds for the Central Universities under Grant No. 2014ZD03-01, and the Program for New Century Excellent Talents in University(NCET-13-0683).

{\small
\bibliographystyle{ieee}
\bibliography{egbib}
}

\end{document}